\documentclass[10pt,twocolumn,letterpaper]{article}

\usepackage{iccv}
\usepackage{times}
\usepackage{epsfig}
\usepackage{graphicx}
\usepackage{amsmath}
\usepackage{amssymb}
\usepackage{multirow}
\usepackage{diagbox}
\usepackage{booktabs}

\usepackage[pagebackref=true,breaklinks=true,letterpaper=true,colorlinks,bookmarks=false]{hyperref}

\iccvfinalcopy 

\def\iccvPaperID{7177} 

\ificcvfinal\fi
\begin{document}
\newcolumntype{L}[1]{>{\raggedright\arraybackslash}p{#1}}
\newcolumntype{C}[1]{>{\centering\arraybackslash}p{#1}}
\newcolumntype{R}[1]{>{\raggedleft\arraybackslash}p{#1}}
\newenvironment{sequation}{\small\begin{equation}}{\end{equation}}
\title{RSC-VAE: Recoding Semantic Consistency Based VAE \\
	   for One-Class Novelty Detection}

\author{Ge Zhang\\
Zhejiang University\\
{\tt\small 21825003@zju.edu.cn}
\and
Wangzhe Du\\
Taiyuan University of Technology\\
{\tt\small duwangzhe@tyut.edu.cn}
}

\maketitle
\ificcvfinal\thispagestyle{empty}\fi
\begin{abstract}
In recent years, there is an increasing interests in reconstruction based generative models for image One-Class Novelty Detection, most of which only focus on image-level information. While in this paper, we further exploit the latent space of Variational Auto-encoder (VAE), a typical reconstruction based model, and we innovatively divide it into three regions: Normal/Anomalous/Unknown-semantic-region. Based on this hypothesis, we propose a new VAE architecture, \textbf{Recoding Semantic Consistency Based VAE (RSC-VAE)}, combining VAE with recoding mechanism and constraining the semantic consistency of two encodings. We come up with three training modes of RSC-VAE: 1. One-Class Training Mode, alleviating False Positive problem of normal samples; 2. Distributionally-Shifted Training Mode, alleviating False Negative problem of anomalous samples; 3. Extremely-Imbalanced Training Mode, introducing a small number of anomalous samples for training to enhance the second mode. The experimental results on multiple datasets demonstrate that our mechanism achieves state-of-the-art performance in various baselines including VAE.
\end{abstract}
\vspace{-0.5cm} 
\section{Introduction}
The imbalance of data distribution is a widespread problem during the deployment of machine learning and deep learning algorithms, where the normal sample always occupies the majority. One-Class Novelty Detection algorithms\cite{ruff2018deep} are very suitable for learning in this scenario since they obtain a certain form of data distribution by only training the normal samples, while the anomalous samples will become outliers during the inference stage because they are out of distribution.

For the past few years, with the constantly development of deep neural networks(DNN)\cite{lecun2015deep}, DNN has been widely used in anomaly detection, where the reconstruction based methods is a significant branch. These model use an encoding-decoding method to reconstruct images, thereby learning the latent space representation of normal training samples\cite{ruff2020unifying}. In the inference stage, the model can reconstruct normal images well, but when inputting anomalous samples, the model will reconstruct as a corresponding normal images or cannot reconstruct the images with anomalous information, that is, it cannot “generalize” the anomalous samples well. In this paper, VAE\cite{kingma2013auto}, a typical reconstruction-based algorithm, is used as the basic structure, whose latent space is deeply explored. The latent space of VAE is firstly divided into three semantic regions: {\bf Normal/Anomalous/Unknown Semantic Regions}, which contain normal samples/anomalous samples/meaningless semantic information, as shown in Figure \ref{fig:4}. Based on this hypothesis, three improved extensions of VAE are proposed.

\begin{figure}[t]
	\begin{center}
		\includegraphics[width=1\linewidth]{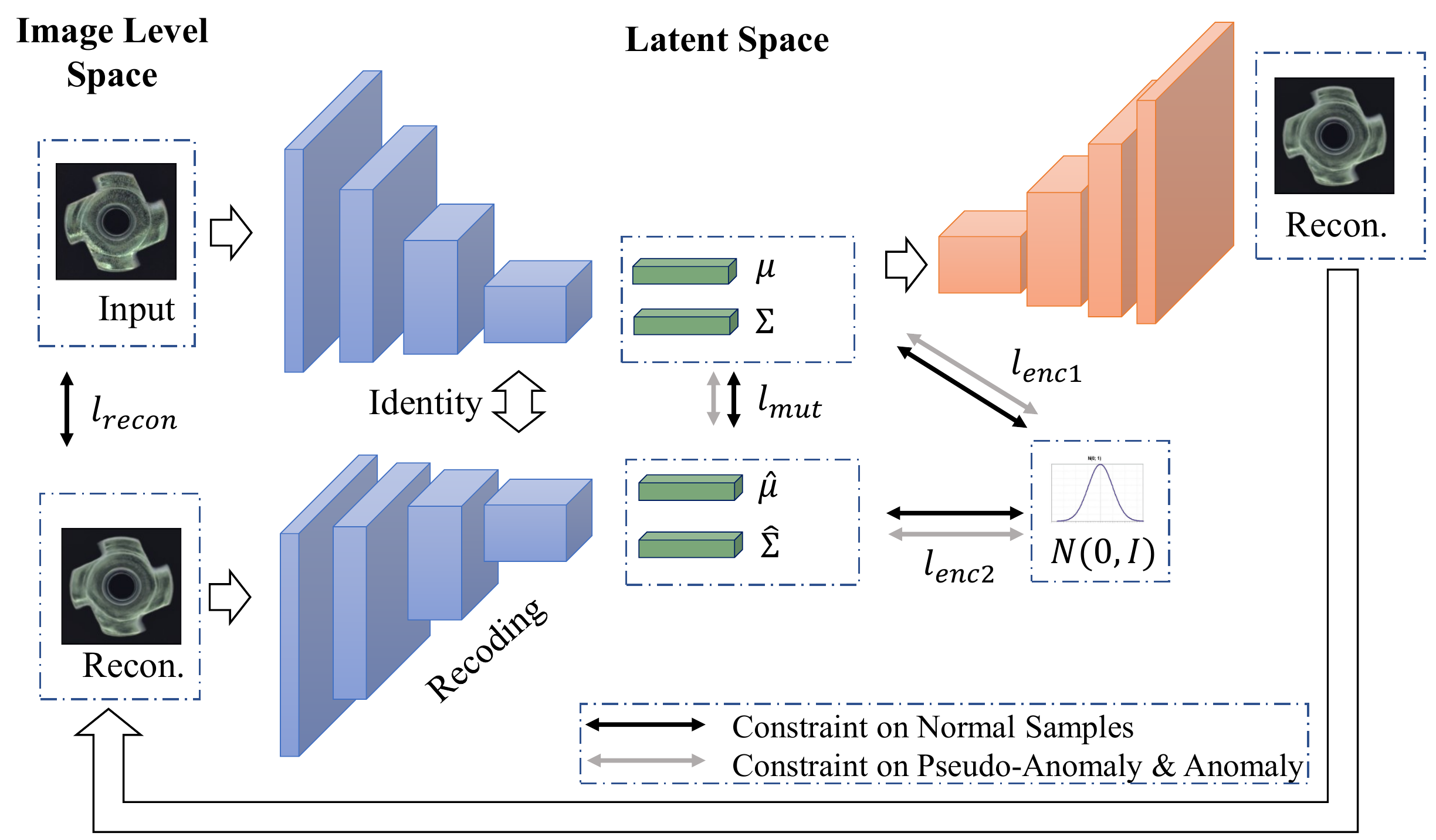}
	\end{center}
	\vspace{-0.1cm} 
	\caption{Illustration of our proposedRSC-VAE: We use VAE as basic architecture and improves its training mechanism. Every input will be encoded twice, first-time encoding and recoding, through one encoder. In RSC-VAE$_o$, only normal images are trained; while in RSC-VAE$_d$/RSC-VAE$_e$, a small number of pseudo anomaly/actual anomaly are introduced for training.}
	\label{fig:1}
	\vspace{-0.5cm} 
\end{figure}

Recently, increasingly reconstruction based methods also regard the difference of encoding between the original image and the reconstruction as a part of the outliers\cite{2018GANomaly}. However, there would be an adverse impact of high non-linearity of neural networks: Even if a normal input is well reconstructed, the encoding of the original image and the reconstruction can be extremely different, shown in Figure \ref{fig:3}(a). As a result, regarding the difference of encoding in the latent space as a part of the outliers may cause False Positive problem of normal samples. In order to alleviate it, we attempt to create a corresponding relationship between image space and Normal-semantic-region in the latent space for normal samples. Therefore, we propose {\bf One-Class Training Mode RSC-VAE (RSC-VAE$_o$)}, an improvement of recoding based VAE. On the basis of the original constraints of VAE—reconstruction consistency and regularization of the first encoding, the proposed RSC-VAE$_o$ recodes the reconstructed image and constrains the consistency of the results of encoding and recoding. In other words, we simultaneously constrain the processes of the encoding-decoding and decoding-encoding.

\begin{figure}[t]
	\begin{center}
		\includegraphics[width=0.75\linewidth]{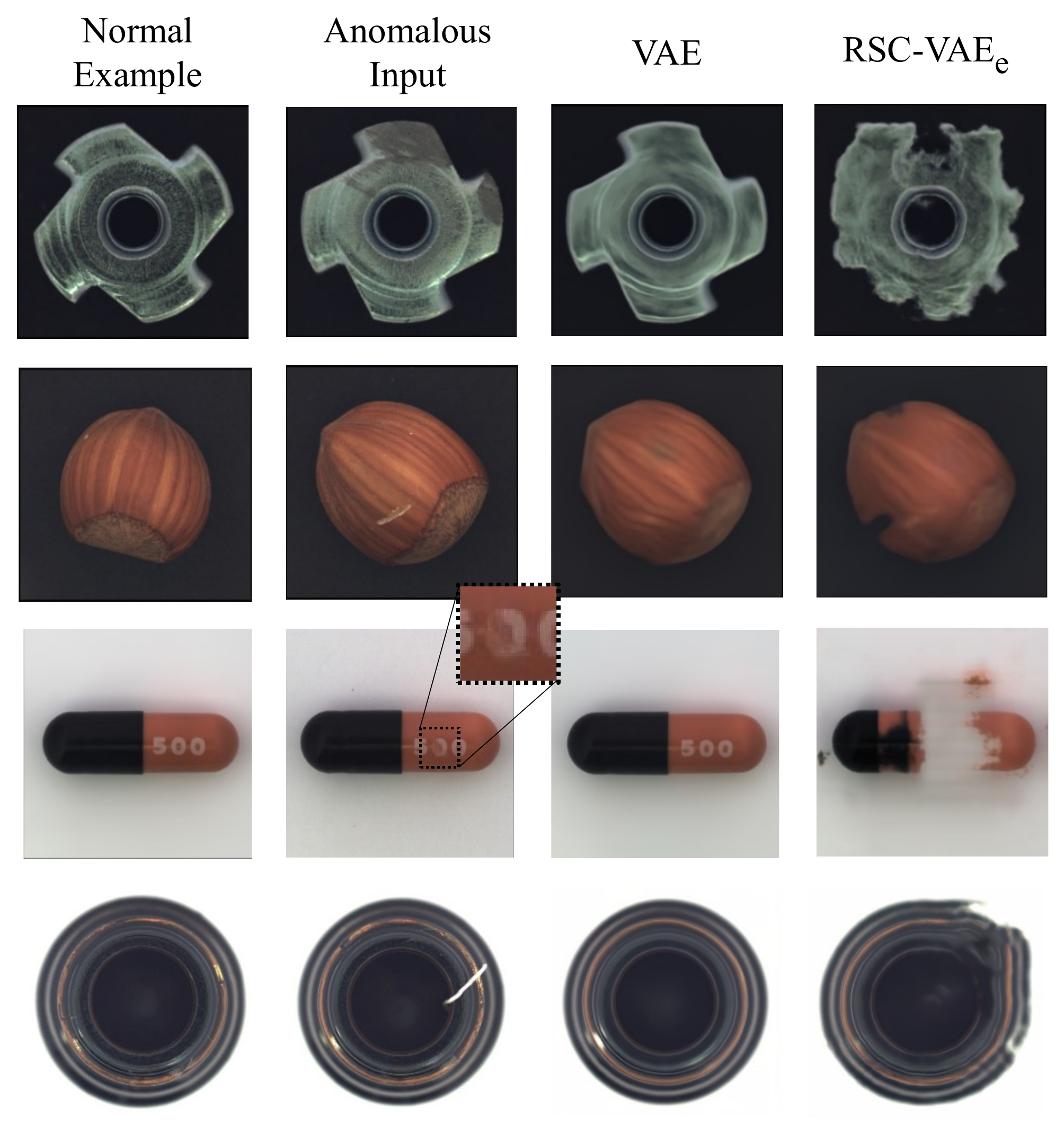}
	\end{center}
	\vspace{-0.2cm} 
	\caption{Comparison of reconstruction between VAE and RSC-VAE$_e$. Traditional reconstruction based scheme tends to ``repair” anomalous regions, resulting in potential False Negative(take the third row for example); While RSC-VAE$_d$/RSC-VAE$_e$ tries to degrade the reconstruction.}
	\label{fig:2}
	\vspace{-0.5cm} 
\end{figure}

In addition, there are two more problems of reconstruction-based anomaly detection, both of which will lead to False Negative problem of anomalous samples: 1. The ability to ``generalize" anomalous inputs is excessive\cite{2020Memorizing}, which means that one anomalous sample can be reconstructed well to obtain lower outliers. 2. The detection effect of images becomes poor when the anomaly of image is not obvious or the anomalous area is small\cite{hong2020latent}. In this case, the reconstruction difference between the reconstructed image and the original image in the anomalous area is very small, resulting in being overwhelmed by the reconstruction error of the normal area (e.g. the third row in Figure \ref{fig:2}).

To alleviate these two problems, we propose {\bf Distributionally-Shifted Training Mode RSC-VAE(RSC-VAE$_d$)}, an extension of RSC-VAE$_o$, which enlarges the reconstruction error by guiding the model to degrade the reconstruction of anomalous inputs. Specifically, a small proportion of inputs are labeled as “pseudo anomaly” by being applied transformation such as rotation, flipping, or cutout, named as ”Distributionally-Shifted Augmentation”(DSA), whose original normal information are losing. Without other constraints, the model cannot control which region (Normal-semantic-region or Anomalous-semantic-region) these pseudo anomalies fall into after being encoded, shown in Figure \ref{fig:4}(a)$\sim$(c)). In this regard, our method is to impose two other constraints on this class of samples in addition to the image consistency constraint in RSC-VAE$_o$: 1. Constrain the result of the first encoding, and guide the feature representation to approach the standard normal distribution. The encoded anomalous samples lose their original semantics and enter into Unknown-semantic-region, resulting in the failure of being decoded into meaningful images. We call this process as “\textbf{Semantic-deviation}”; 2. Constrain the consistency of the first encoding and recoding in the latent space, so as to provide additional guarantees for reducing the model's ability to reconstruct anomalous inputs.

Meanwhile, we noticed that in application domains, apart from a large number of normal samples, there are a small number of anomalous samples labeled by experts in the field. Therefore, in the proposed {\bf Extremely-Imbalanced Training Mode RSC-VAE(RSC-VA$_e$)}, we introduce these precious samples to replace the pseudo anomalies in RSC-VAE$_d$, for the purpose of improving model’s Semantic-deviation capability with regard to actual anomalous information during the training stage, thereby further reducing the model's ability to reconstruct anomalies in the test set.

Here, we summarize the main contributions of this paper:
\begin{itemize}
	\vspace{-0.1cm} 
	\item [(1)]
	Based on VAE, the typical reconstruction based scheme for One-Class Novelty Detection, we deeply explore the value of latent space and divide it into three different regions: Normal/Anomalous/Unknown-semantic-region.
	\vspace{-0.1cm} 
	\item [(2)]
	We propose Recoding Semantic Consistency Based VAE (RSC-VAE) , which combines VAE with the recoding mechanism to constrain the consistency of the semantics of the two encodings, and propose three different training modes: RSC-VAE$_o$, RSC-VAE$_d$, and RSC-VAE$_e$. The first mode alleviates the False Positive problem of normal samples, the latter two modes alleviate the False Negative problem of anomalous samples by using our proposed 
	 ``Semantic-deviation" mechanism, and the third mode enhances model's ability to explore the real anomalous information;
	\vspace{-0.1cm} 
	\item [(3)]Our experimental results on MNIST, FashionMNIST, and MVTEC datasets demonstrate that the proposed mechanisms effectively improve the performance of VAE and surpass the performance on various state-of-the-art baselines.
	
\end{itemize}

\section{Related Work}

\noindent{\bf Recoding.}
Recoding refers to a mechanism that implement a second encoding, to the reconstructed image based on encoder-decoder architectures such as Autoencoder(AE), Variational Autoencoder (VAE)\cite{kingma2013auto}, and Denoising Autoencoder, which has been broadly used in various tasks in recent years. In the image generation task, for example, \cite{hou2017deep} maps the original image and the reconstructed image to a new encoder and constrains the consistency, aiming to improve the image quality generated by the decoder; \cite{jha2018disentangling} splits the encoding results of VAE into two branches, and disentangles the feature representation based on recoding mechanism.

In this paper, the recoding mechanism is utilized in the anomaly detection task so as to take the difference between the encoding results of the original image and the reconstruction as a part of the outliers, for the purpose of better mining the anomalous information. The architecture of AnoGAN\cite{schlegl2017unsupervised} is exactly the same as GAN\cite{goodfellow2014generative} in the training process. But in inference stage, an objective function is used to update the representation in latent space through back-propagation, which dramatically reduces the inference efficiency of the model. Ganomaly\cite{2018GANomaly} recodes the reconstructed image based on AAE\cite{makhzani2015adversarial} architecture, but a new encoder is employed for recoding, so that the two features are not in the same latent space, which reduces the value of feature differences. In contrast, encodings in this paper use the same encoder. Besides, both \cite{kim2019rapp} and \cite{komoto2020consistency} do use only one encoder for recoding in the inference stage, but not perform the same consistency constraint as we do in the training stage.

\noindent{\bf Distributionally-shifted image transformation.}
Different from the commonly used data augmentation, a lot of work has involved pretext tasks\cite{he2020momentum}, using data preprocessing to change or erase its original information, which is known as distributedally-shifted image transformation\cite{tack2020csi}. For example, contrastive learning\cite{he2020momentum, chen2020simple} applies the instance-level recognition to pre-train the model based on pretext task, so that the skeleton network can better represent the input. In the anomaly detection task, GeoTrans\cite{golan2018deep} applies a variety of geometric transformations and takes them as classification label for training. In the inference stage, it gives samples with low classification probabilities high anomaly scores. Based on this mechanism, \cite{bergman2020classification} introduces the distance measurement into the loss function and the measure of the outliers. On the basis of Autoencoder, \cite{fei2020attribute}, \cite{ulutas2020split} and \cite{zhao2018surface} restore images whose information have been erased due to the preprocessing (including rotation, removal of channels, cutout\cite{devries2017improved}, etc.). CSI\cite{tack2020csi} introduces contrastive learning into anomaly detection, uses distributedally-shifted image as a negative sample to alleviate the performance penalty problem mentioned in \cite{chen2020simple}. In this paper, we also take advantages of distributionally-shifted image transformation to forge a small number of anomalous samples, however, different from the pretext task described above, our purpose is to guide the model convert the representations that originally fallen in Anomalous-semantic-region to Unknown-semantic-region.

\noindent{\bf One-class classification with Several anomalous samples.}
In practical applications, apart from a large number of normal samples, there are also a small number of anomalous samples labeled by experts in the field. Some researchers adopt them in the training stage of original One-Class Novelty Detection to achieve better performance. For instance, the algorithm CAVGA\cite{venkataramanan2019attention} uses both unsupervised and semi-supervised strategies, the latter adopt Grad-CAM\cite{selvaraju2017grad} to generate an anomalous attention map and calculates the complementary guided attention loss to minimize anomalous attention that obtained the model normal semantic information. FCDD\cite{liznerski2020explainable} also proposes the semi-supervised FCDD on the basis of the original unsupervised architecture, which only needs to modify the loss function. However, the ground-truth anomalous mask region of the sample is required in both two algorithms above. Our work also makes use of a small number of anomalous samples to construct a variant of the One-class classification task, which only needs to modify the loss function.

\section{Methodology}
In this section, we will comprehensively describe the proposed RSC-VAE. The overall architecture of RSC-VAE is shown in Figure \ref{fig:1}, with three training modes: One-Class Training Mode / Distributionally-Shifted Training Mode / Extremely-Imbalanced Training Mode, denoted as RSC-VAE$_o$, RSC-VAE$_d$, RSC-VAE$_e$, respectively. The model abandons the widely used adversarial training strategy, only uses the conventional VAE structure with one encoder-decoder and improves its training mechanism. In Section 3.1, we will review VAE, and then introduce the three modes respectively.

\subsection{Basic Architecture: VAE}

We use VAE\cite{kingma2013auto} as basic architecture, and reform its training mechanism. As with all generative models, VAE aims to obtain the posterior probability distribution $p(z | x)$:
\vspace{-0.18cm}
\begin{sequation}
	p(z|x) = \frac{p(x|z)p(z)}{\int p(x|z) p(z) dz}
	\vspace{-0.18cm}
\end{sequation}

VAE introduces $q(z | x)$ to perform variational inference instead of using Bayes Formula which has the problem of intractability. That is, minimizing the $KL$ divergence\cite{kullback1951} of two distributions.
\vspace{-0.18cm}
\begin{sequation}
	\min D_{KL}(q(z|x)||p(z|x))
	\vspace{-0.3cm}
\end{sequation}

This objective function is equivalent to maximizing the lower bound of the variable fraction $\mathcal{L}_b$
\begin{sequation}
	\begin{split}
		&\max \mathcal{L}_b = \max \int q(z|x) \log \frac{p(x,z)}{q(z|x)} dz\\
		=\max&\left[ \int q(z|x)\log p(x|z)dz + \int q(z|x)\log \frac{p(z)}{q(z|x)}dz \right]\\
		=\max &\left[\mathbb{E}_{q(z|x)} [\log p(x|z)] - D_{KL}[q(z|x) || p(z)] \right]
		\label{vae-formula}
	\end{split}
\end{sequation}
\noindent{where $p(z)=N(0,I)$.}

The reason why we choose VAE as the basic architecture is that VAE has its own regularization for its latent space. On the one hand, we can better explore the latent space of Autoencoder architecture under the benefit of the “continuous and complete” characteristic of the one in VAE; On the other hand, we can directly use this regularization term to perform Semantic-deviation for pseudo anomalies/anomalous samples in RSC-VAE$_d$/RSC-VAE$_e$, without establishing other constraints.

\subsection{One-class Training Mode}

\begin{figure}[t]
	\begin{center}
		\includegraphics[width=0.8\linewidth]{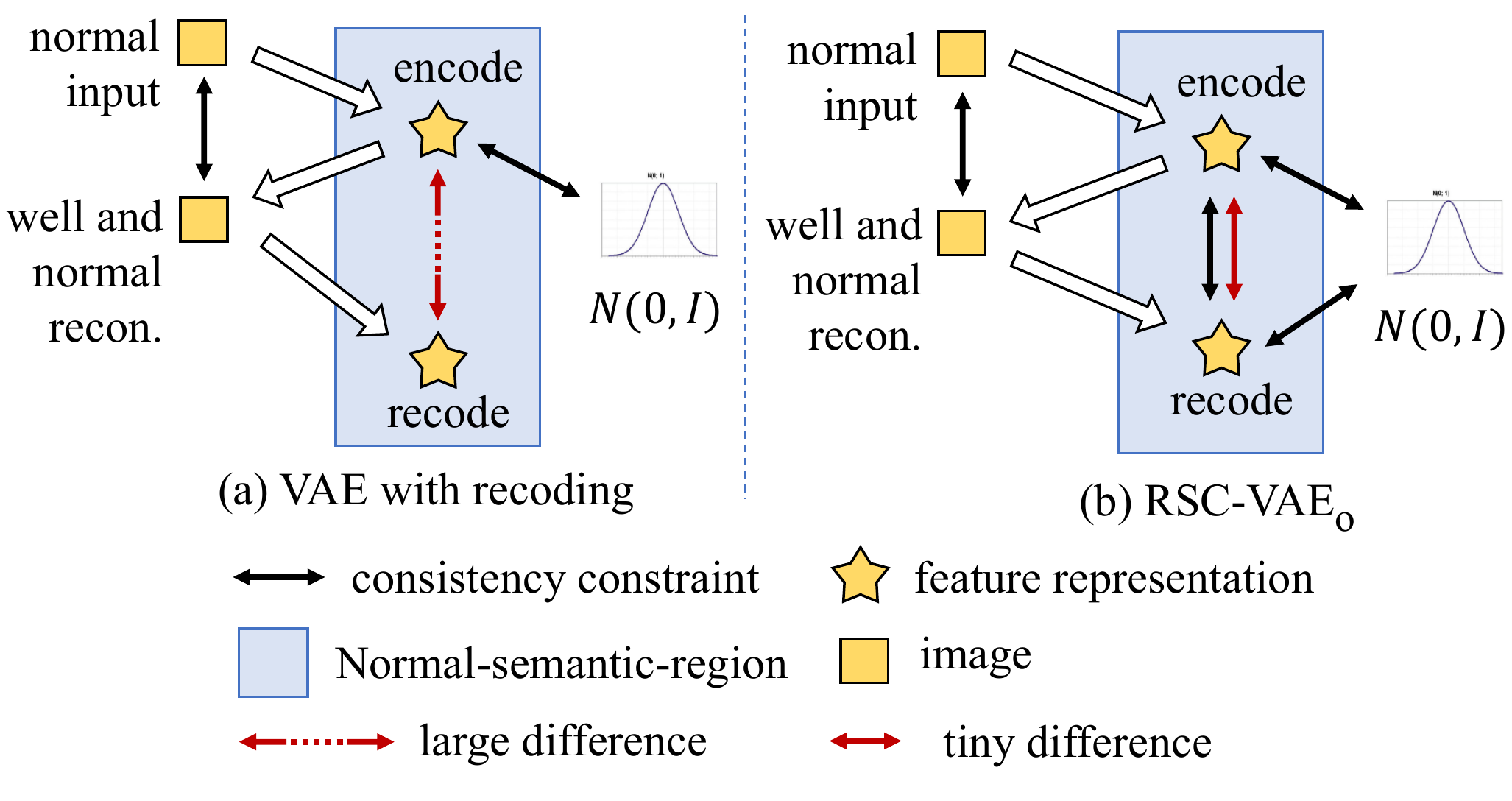}
	\end{center}
	\vspace{-0.2cm} 
	\caption{RSC-VAE$_o$’s solution for False Positive problem in recoding based VAE.}
	\label{fig:3}
	\vspace{-0.5cm} 
\end{figure}

Noted that a growing amount of works uses the weighted sum of image-level reconstruction errors with the difference between two encodings’ results as final outliers. Nevertheless, there would be a problem of it: Many anti-attack algorithms have found that the fine-tuning of images which are almost invisible to the naked eye can also make huge changes in the results of the feature space\cite{8294186}. Videlicet, even if a normal input is well reconstructed, the encoding of the original and reconstructed images may differ greatly, shown in Figure \ref{fig:3}(a). On this ground, taking the difference of latent space as a part of the outliers may result in False Positive problem for normal samples.

In response to this problem, we first propose a definition: In a VAE based One-Class Novelty Detection task, the high-dimensional manifold region fitted by the feature representations of all normal samples is \textbf{Normal-semantic-region}, the distribution of which covers complete information of normal samples. In order to alleviate the False Positive problem above, we reckon that the image space of normal samples and the Normal-semantic-region should have a corresponding relationship: Two feature representations, who located in Normal-semantic-region and represent two similar images, should be close. Hence, we improve the mechanism of recoding based VAE and propose One-Class Training Mode RSC-VAE (RSC-VAE$_o$) to realize the One-Class Novel Detection task. For enhancing this corresponding relationship, RSC-VAE$_o$ combines the consistency constraint of original image-level reconstruction with that of the recoding results and the first encoding results, shown in Figure \ref{fig:3}(b), which mitigating the distortion of normal information due to transmission between two spaces. The objective function for normal inputs will be clarified below.

\noindent{\bf 1. VAE Loss}

Since the first encoding and decoding process is consistent with conventional VAE, in this part we continue to use the loss function of VAE (Eq. \ref{vae-formula}). In the formula above, the first term can be considered as a reconstruction loss. In this paper, we choose $L_1$ loss as its implementation. We denote the encoder be $F$ and the decoder be $G$. $z=\{\mu,\Sigma\}=F(x)$ calculates the parameters of the posterior probability distribution $q(z|x)\sim N(\mu,\Sigma)$ of VAE, where $\mu_{D \times 1}$ and $\Sigma_{D \times D}$ represent the mean vector and covariance matrix of the normal distribution respectively and $D$ represents the dimensions of the latent space. $\hat{x} = G(z)$ calculates the likelihood $p(x|z)$ of $x$ with regard to $z$. $\hat{z} = \{ \hat{\mu}, \hat{\Sigma} \} = F(\hat{x})$ calculates the parameter of posterior probability distribution $q(\hat{z}|\hat{x})\sim N(\hat{\mu}, \hat{\Sigma})$ of recoding.
We rewrite the objective function of VAE as
\vspace{-0.1cm}
\begin{sequation}
	l_{recon}(x) = ||x - \hat{x}||_1
\end{sequation}
\vspace{-0.1cm}
\begin{sequation}
		l_{enc1}(x) = D_{KL} [q(z|x) || p(z)] = D_{KL} [ N(\mu, \Sigma) || N(0,I)]\\
\end{sequation}

\begin{sequation}
	l_{vae}(x) = l_{recon}(x) + l_{enc1}(x)
\vspace{-0.4cm}
\end{sequation}

Note that due to the existence of $l_{enc1}$, the Normal-semantic-region will drift towards a standard normal distribution with the same dimensions, and the reconstruction constraint $l_{recon}$ is utilized to maintain the generation ability of the decoder. These two items restrict each other so that VAE’s latent space has continous semantic and maintains complete normal information.

\noindent{\bf 2. Consistency of Latent Space}

RSC-VAE$_o$ hopes that the result of recoding and the first encoding tend to have the same distribution. Firstly, since the regularization of the first encoding is imposed, the same regularization should be imposed to the distribution $ N(\hat{\mu}, \hat{\sigma})$ from recoding as well.
\vspace{-0.1cm}
\begin{sequation}
		l_{enc2}(x) = D_{KL}[q(\hat{z} | \hat{x}) || p(\hat{z})] = D_{KL}[N(\hat{\mu}, \hat{\Sigma}) || N(0,I)] \\
	\vspace{-0.4cm}
\end{sequation}

Secondly, because $l_{enc2}$ is not sufficient to ensure the consistency of two representations, a consistency constraint must be defined explicitly. Under the VAE mechanism, we constrain the parameters of two normal distributions $ N(\mu, \Sigma)$ and $N(\hat{\mu}, \hat{\Sigma})$.

\begin{sequation}
l_{mut}(x) = D_{JS}[q(z|x) || q(\hat{z} | \hat{x}) ]
\vspace{-0.3cm}
\end{sequation}

Finally, we let the latent space consistency constraint as

\begin{sequation}
	l_{consist}(x) = l_{enc2}(x) + l_{mut}(x)
	\vspace{-0.3cm}
\end{sequation}

\noindent{\bf 3.Total loss}

The total loss is the weighted sum of two losses above. Suppose there are $N$ samples in one batch $\{ x_i \}_{i=1}^N$, we have
\vspace{-0.3cm}
\begin{sequation}
	L_o = \sum_{i=1}^{N}l_{normal}(x_i) = \sum_{i=1}^{N}[l_{vae}(x_i) + \lambda l_{consist}(x_i)]
	\vspace{-0.2cm}
\end{sequation}

\noindent where $\lambda$ is a hyper-parameter which is fixed at 0.1.

\subsection{Three possible results after inputting anomalous samples}
\begin{figure}[t]
	\begin{center}
		\includegraphics[width=0.9\linewidth]{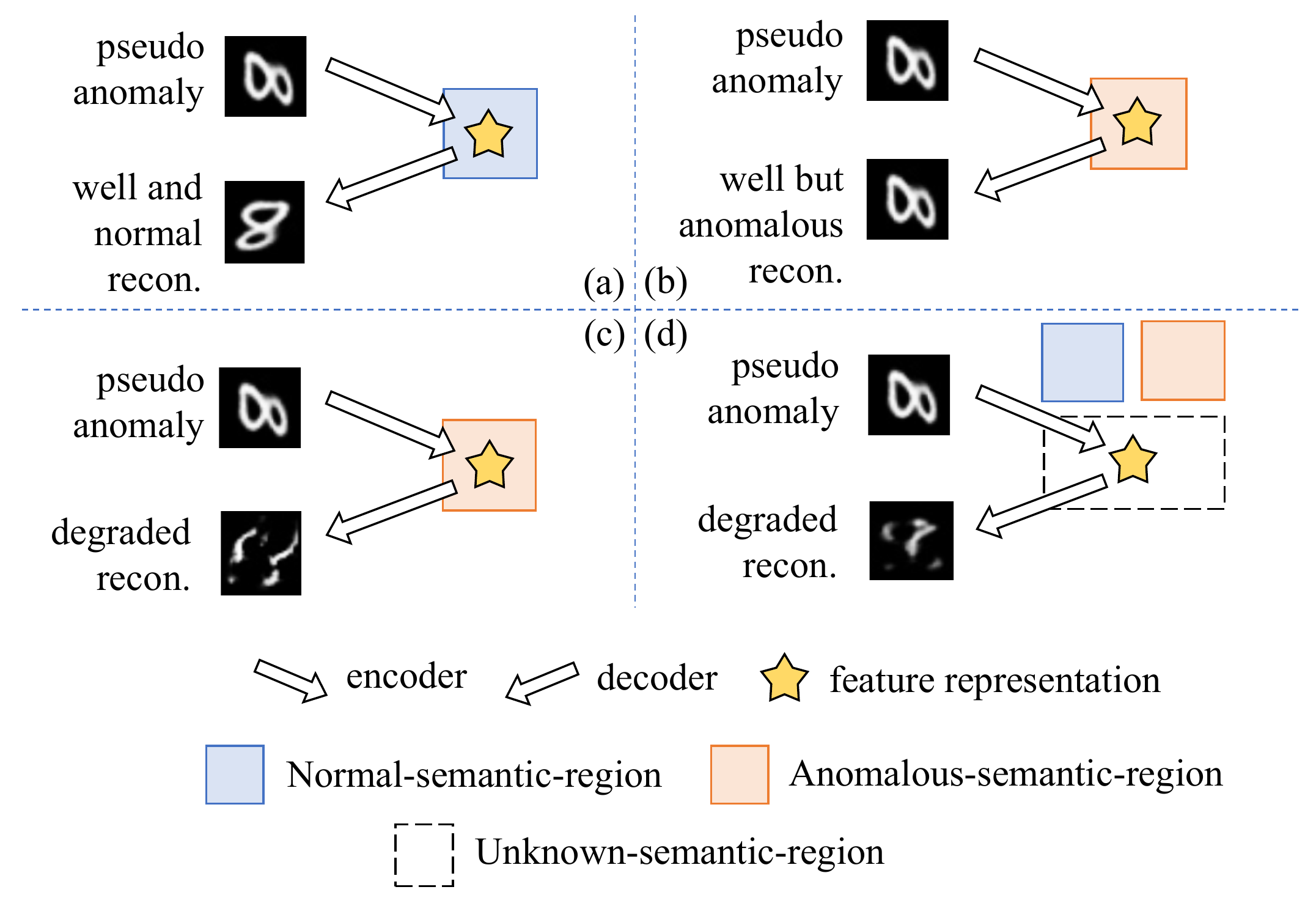}
	\end{center}
	\vspace{-0.2cm} 
	\caption{(a)$\sim$(c) Three different encoding-decoding process in VAE when considering pseudo anomalies as inputs. (d) The process under the mechanism of our proposed RSC-VAE$_d$. In this diagram, digit ``8” is normal category and the setting of DSA is rotation.}
	\label{fig:4}
	\vspace{-0.5cm} 
\end{figure}

Before discussing RSC-VAE$_d$, we first analyze the encoding and decoding results of VAE for anomalous input. As shown in Figure \ref{fig:4}, we adopt a baseline VAE which has been trained under One-Class Novelty Detection Task (assuming we take digit “8” for training). If an anomalous sample (the number ``8" rotated by 90°) is input, the encoding and the decoding result can be classified into three cases.

\noindent{\bf Case 1.} After the anomalous input is encoded, its feature representation falls into Normal-semantic-region, and is subsequently decoded to generate the corresponding normal samples. Shown in Figure \ref{fig:4}(a).

\noindent{\bf Case 2.} The anomalous input can be well reconstructed; hence we believe the representation of this sample in the latent space should have complete anomaly information. Here, we define the high-dimensional manifold fitted by all the representations that are capable of storing complete anomaly information and may exist decoders for complete reconstruction of them as \textbf{Anomalous-semantic-region}. Shown in Figure \ref{fig:4}(b).

\noindent{\bf Case 3.} The anomalous input enters Anomalous-semantic-region after been encoded, but the decoder is only expert in processing the normal semantic information, resulting in a chaotic/degrade reconstruction. Shown in Figure \ref{fig:4}(c).

In contrast to case 3, the decoder in case 2 has the ability to ``generalize" the anomalous semantics, and therefore is able to reconstruct the anomalous samples, giving us the unwanted False Negative problem. The reason for case 1 is that, compared to the latter two cases, only the encoder in case 1 is able to gradually filter the anomalous information during the encoding process, leaving only the normal information.

Numerous existing methods\cite{2020Memorizing, hong2020latent} try to guide the anomalous samples into case 1 by enhancing the regularity or sparsity of the latent space. But case 1 may also have another False Negative problem: If the anomaly of image is not obvious or the anomalous area is small, the reconstruction difference between the reconstructed image and the original image in the anomalous area is very small, resulting in being overwhelmed by the reconstruction error of the normal area.

\begin{figure}[t]
	\begin{center}
		\includegraphics[width=0.65\linewidth]{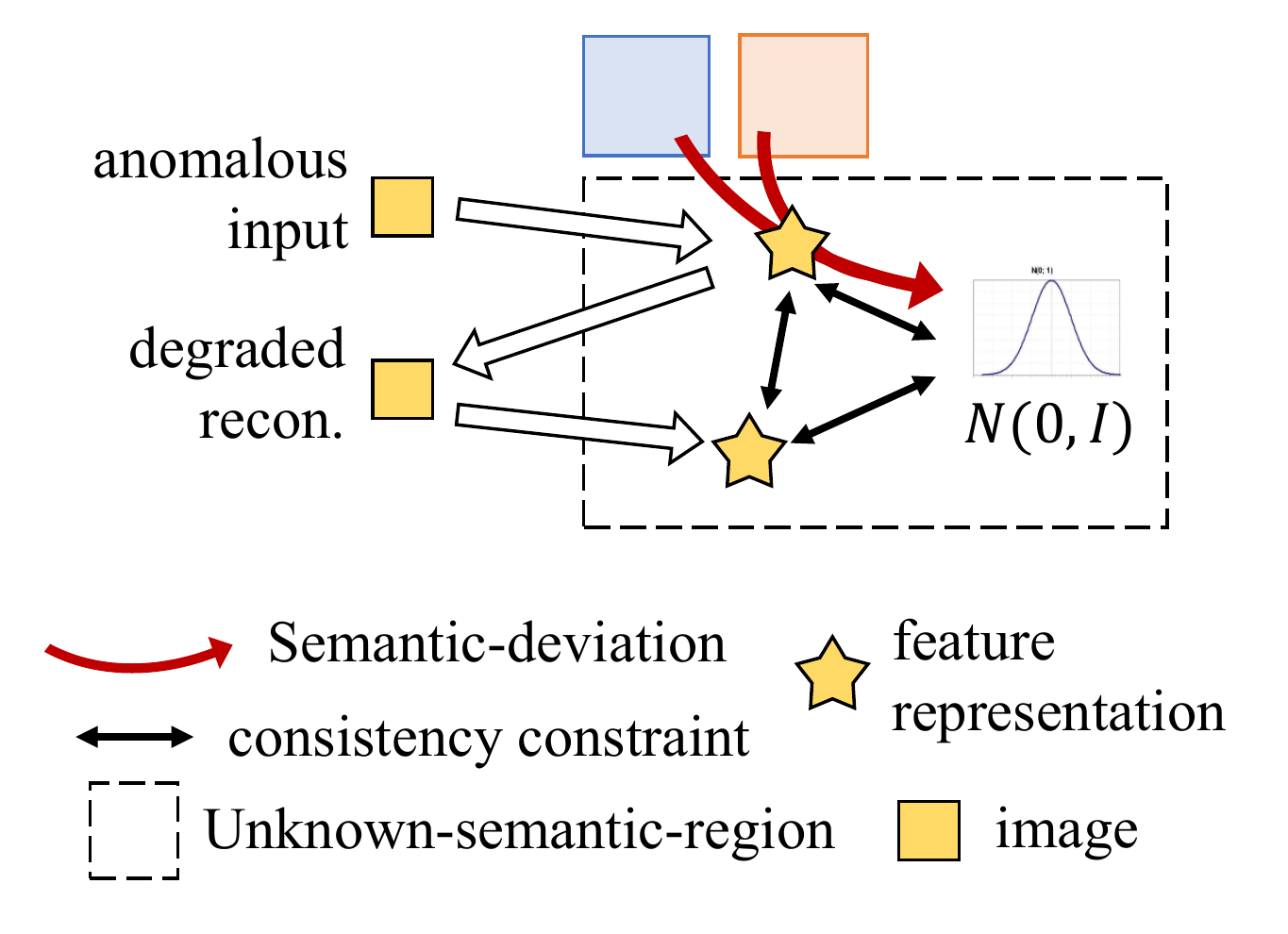}
	\end{center}
	\vspace{-0.3cm} 
	\caption{Illustration of Semantic-deviation and constraint in RSC-VAE$_d$/RSC-VAE$_e$.}
	\label{fig:5}
	\vspace{-0.5cm} 
\end{figure}

\subsection{Distributionally-shifted Training Mode}
RSC-VAE$_o$ mitigates the False Positive problem of normal input under the recoding mechanism, but cannot solve the False Negative problem of anomalous input as described above. Hence, in this section, we extend the RSC-VAE$_o$ algorithm and propose a Distributionally-Shifted Training Mode RSC-VAE (RSC-VAE$_d$) . The motivation of this mode is to make sure that the feature representations of the anomalous samples neither fall into the Normal-semantic-region, which prevents case 1, nor fall into the Anomalous-semantic-region, which prevents case 2 and 3. Here, we define the region that does not belong to the former semantic region as \textbf{Unknown-semantic-region}, shown in Figure \ref{fig:4}(d). The feature representations in this region have neither one normal image’s complete information nor one anomalous image’s complete information, which results in a chaotic image after being reconstructed.

RSC-VAE$_d$ is still a One-Class Novelty Detection task, but a proportion of training samples are applied pre-processing such as rotation/flip/cutout\cite{devries2017improved}, which will erase some important information of normal sample and deviates away from the distribution of original normal sample. We call this pre-process as {\bf Distributedally-Shifted Augmentation (DSA)} , and call these samples as ``{\bf pseudo anomaly}". Unlike data augmentation such as adding noise or changing contrast, DSA dramatically changes the feature representation of one image (SimCLR\cite{chen2020simple} demonstrates its negative effects on contrastive learning task), hence the label can be transferred to anomalous under RSC-VAE$_d$, which would not happen in traditional data augmentation.

The loss function of RSC-VAE$_d$ is explained below. This training mode cancels the reconstruction constraint $l_{recon}$, but still performs the regularization $l_{enc1}$ and the consistency constraint $l_{consist}$. The purpose of not constraining $l_{recon}$ is to prevent VAE from having reconstruction capability for pseudo anomaly. The mechanism of using $l_{enc1}$ and $l_{consist}$ here is different from the one in RSC-VAE$_o$:

\noindent{\bf Mechanism of $l_{enc1}$:} We hope to make use of this constraint to guide the representations that ought to fall into the Anomalous-semantic-region or the Normal-semantic-region drift towards the standard normal distribution, thereby entering into the Unknown-semantic-region and decoded as a chaotic image, shown in Figure \ref{fig:5}. The reason for using the VAE architecture is demonstrated again: the regularization term of the VAE can be used directly to applying Semantic-deviation for pseudo anomaly.

\noindent{\bf Mechanism of $l_{consist}$:} This constraint aims to further reduce the model’s reconstruction ability for anomalous samples. Since there is no reconstruction constraint $l_{recon}$, the model does not require the reconstruction quality of pseudo anomalies. According to the Maximum Entropy Principle\cite{6773024}, among all possible probabilistic models, the model with the highest entropy is the best one, so the semantic of the image space further decreases in the decoding -recoding stage and can be reduced to a Gaussian Distribution theoretically (the probability distribution with the highest information entropy is a Gaussian Distribution). From another perspective, the decoding-recoding process can be considered as an Autoencoder whose latent space is a visual-level space, imposing $l_{consist}$ increases the sparsity of this high-dimensional space, which means the generation of incomplete information.

We denote DSA as $ds(x)$, the loss function of a sample that is applied by DSA is
\begin{equation}
	l_{pseudo}(x) = l_{enc1}(ds(x)) + \lambda l_{consist}(ds(x))
\end{equation}

Let $\{ y_i \}_{i=1}^N$ as a group of samples in the Bernoulli distribution $Y \sim B(N,p)$, denoting whether or not applying DSA for the samples in the batch. We will assign a weight $\beta$ to these small number of pseudo anomalies depending on the dataset. Hence the final loss function of a batch is
\vspace{-0.3cm}
\begin{sequation}
	L_d = \sum_{i=1}^{N} [(1 - y_i) \cdot l_{normal}(x_i) + \beta \cdot y_i \cdot l_{pseudo}(x_i)]
	\vspace{-0.5cm}
\end{sequation}

\subsection{Extremely-imbalanced Training Mode}
In the Extremely-Imbalanced Training Mode RSC-VAE(RSC-VAE$_e$) , we introduce anomalous samples marked by experts in the field to replace the pseudo anomaly samples in RSC-VAE$_d$. In RSC-VAE$_d$, we treat the image pre-processing as anomalous information, but the corresponding Anomalous-semantic-region may not perfectly represent the anomalous samples in the test set (though there has been improvement, shown in Experiments). Hence in the mode RSC-VAE$_e$, which is trained with actual anomalies, the model obtains better ability of Semantic-deviation and lower ability of reconstruction for the representations of actual anomaly information that are located in the Anomalous-semantic-region. We denote $\{ y_i \}_{i=1}^N$ as the label of one batch, $y_i=0$ denotes normal, $y_i=1$ denotes anomalous. Hence the loss function is 

\vspace{-0.1cm}
\begin{sequation}
	l_{anomalous}(x) = l_{enc1}(x) + \lambda l_{consist}(x)
	\vspace{-0.4cm}
\end{sequation}
\begin{sequation}
	L_e = \sum_{i=1}^{N} [(1 - y_i) \cdot l_{normal}(x_i) + \beta \cdot y_i \cdot l_{anomalous}(x_i)]
\end{sequation}

\subsection{Inference Stage}

In inference stage, the measure of outlier uses the weighted sum of the reconstruction error from the image level and the feature level. Denote $x_i$, $i = 1,2,...,n$ as $n$ test samples, we calculate the outlier $s_i$ for each sample.
\vspace{-0.2cm}
\begin{sequation}
	s_i = \alpha \cdot \frac{l_{mut}(x_i)}{\mathbb{E}_x[l_{mut}(x)]} + (1 - \alpha) \cdot \frac{l_{recon}(x_i)}{\mathbb{E}_x[l_{recon}(x)]}
	\label{abn-formula}
\end{sequation}
\vspace{-0.2cm}

After calculating the outlier $s_i$ for all test samples, we perform normalization to obtain the final outlier $s_i^\prime$
\vspace{-0.1cm}
\begin{sequation}
	s_i^\prime = \frac{s_i - \underset{i}{\min}(s_i)}{\underset{i}{\max}(s_i) - \underset{i}{\min}(s_i)}
	\vspace{-0.3cm}
\end{sequation}

Subsequently, the evaluation metrics are calculated based on the ground-truth label.

\section{Experiments}
In this section, we employ Area Under the Curve (AUC) to evaluate our framework in 30 categories from three datasets. Different types and aspect of experiments will be conducted to verify the effectiveness of all the three training models we proposed.

\subsection{Dataset and Training Setup}

\noindent{\bf a) MNIST and fMNIST}

Each of MNIST\cite{lecun1998mnist} and fMNIST\cite{xiao2017fashion} consists of 60,000 training samples and 10,000 test samples, but needs to be re-divided in One-Class Classification scenario. Ganomaly\cite{2018GANomaly}’s division scheme is referenced in this paper. Specifically, we first specify one target category as normal category, the remaining as anomalous. The training set is made up of all target class samples in the original training set, while the normal sample in the testing set consists of all target class samples in the original testing set and the anomalous sample contains 50\% of all non-target class samples in the original testing set. Since the dataset contains ten numbers, a total of ten training-test set combinations are partitioned.

When training under RSC-VAE$_d$, the probability of DSA is fixed at 1\%, and 1$\sim$3 options includes horizontal flip/vertical flip/rotate 90° are randomly selected; While in RSC-VAE$_e$, a certain percentage of anomalous samples are added to the original training set to assist the training.

For image preprocessing, the image size is adjusted from 28$\times$28 to 32$\times$32 without any other data augmentation except for normalization. As for the encoder structure, the encoder is composed of four 3$\times$3 convolution which are connected in series, each followed by a Batch Normalization layer and a LeakyReLU activation function layer (slope = 0.2). The model is trained and tested based on PyTorch\cite{paszke2017automatic} by optimizing the networks using Adam with an initial learning rate set to 0.01 and updating through the cosine annealing strategy where $T_{max}$ set to 50. The total training epoches is set to 500 and the batchsize is set to 32. The loss function tends to flatten out after a certain epoch of model training, and we select the optimal AUROC in the test set after the loss function converges. In RSC-VAE$_d$ and RSC-VAE$_e$, we set $\beta$=1 in the anomaly sample loss function, while in the inference stage, we set $\alpha$=0.5 in Eq.\ref{abn-formula}.

\noindent{\bf b) MVTecAD}

MVTecAD\cite{bergmann2019mvtec} is an industrial dataset containing more than 5,000 high-resolution images, with 5 texture classes and 10 object classes. Each category contains both normal images for training or testing, and images with various types of anomalies for testing.

In RSC-VAE$_d$, we choose random cutout\cite{devries2017improved} as DSA since this processing is more similar with the characteristics of anomalous images and the probability of it is fixed at 5\%, while in RSC-VAE$_e$, we randomly select a fixed number of samples in anomalous subclasses in each category of MVTecAD to assist training. Samples that appear in the training set will not appear in the test set to avoid data leakage.

The image size is adjusted from original 1024$\times$1024 to 256$\times$256 in pre-processing. Before the images are fed into the network, random data augmentation is performed according to its category, including horizontal flip, vertical flip, and slightly rigid transformation. For the sake of reducing parameters and preventing overfitting due to the small magnitude of the training sets, the encoder structure in this paper adopts Resnet-18 as the skeleton network where the channel width of all layers are fixed to 64. Unlike the parameter settings in MNIST/fMNIST, the batch size is set to 8. In addition, due to the small magnitude of MVTecAD data, the metric is susceptible to fluctuate, so in RSC-VAE$_d$ and RSC-VAE$_e$, we trained each class at the settings of $\beta$=1,5,10, respectively, and selected the optimal results.

\begin{table}
	\newcommand{\tabincell}[2]{\begin{tabular}{@{}#1@{}}#2\end{tabular}}  
	\scriptsize
	\begin{center}
		\linespread{1.05}\selectfont
		\begin{tabular}{C{7mm}|L{5.1mm}L{5.1mm}L{5.1mm}L{5.1mm}|L{5.1mm}L{5.1mm}L{5.1mm}L{5.1mm}}
			\hline\hline
			Category &	AE &	Gano-maly  &	OC-GAN &	Grad-Con & Recode VAE &	RSC-VAE$_o$ &	RSC-VAE$_d$ & RSC-VAE$_e$ \\
			\hline
			0 &	0.968 &	0.972  & \underline{0.998} & 0.995 & 0.996 & \bf{0.999} &	0.997 &	\bf{0.999} \\
			1 &	\underline{0.997} &	0.996  & \bf{0.999} & \bf{0.999} & \bf{0.999} & \bf{0.999} & \bf{0.999} & \bf{0.999} \\
			2 &	0.882 &	0.851  &	0.947 &	0.952 &	0.958 &	0.972 & \underline{0.978} &	\bf{0.999} \\
			3 &	0.905 &	0.906  &	0.962 &	\underline{0.973} &	0.957 &	0.942 &	0.963 &	\bf{0.999} \\
			4 &	0.917 &	0.949  &	0.972 &	0.969 &	0.972 &	0.978 &	\underline{0.984} &	\bf{0.999} \\
			5 &	0.907 &	0.949  &	0.980 &	0.977 &	0.974 &	\underline{0.982} &	0.977 &	\bf{0.999} \\
			6 &	0.959 &	0.971  &	0.991 &	0.994 &	0.996 &	\underline{0.998} &	0.996 &	\bf{0.999} \\
			7 &	0.964 &	0.939  &	0.980 &	0.979 &	0.970 &	0.971 &	\underline{0.983} &	\bf{0.999} \\
			8 &	0.821 &	0.797  &	0.938 &	0.919 &	\underline{0.957} &	0.941 &	0.942 &	\bf{0.998} \\
			9 &	0.947 &	0.954  &	0.981 &	0.973 &	0.983 &	0.981 &	\underline{0.988} &	\bf{0.998} \\
			\hline
			Avg. &	0.928 &	0.928  &	0.975 &	0.973 &	0.976 &	0.976 & \underline{0.980} &	\bf{0.998} \\
			\hline\hline
		\end{tabular}
	\end{center}
	\vspace{-0.25cm} 
	\caption{Results of RSC-VAE in MNIST dataset.\label{tab:1}}
\end{table}

\begin{table}
	\scriptsize
	\begin{center}
		\linespread{1.05}\selectfont
		\begin{tabular}{L{7mm}|L{5.1mm}L{5.1mm}L{5.1mm}L{5.1mm}|L{5.1mm}L{5.1mm}L{5.1mm}L{5.1mm}}
			\hline\hline
			Category &	AE &	Gano-maly &	AD-GAN &	DSE-BM &	Recode VAE &	RSC-VAE$_o$ &	RSC-VAE$_d$ &	RSC-VAE$_e$ \\
			\hline
			T-shirt &	0.700 &	0.693 &	0.890 &	0.860 &	0.893 &	0.904 &	\underline{0.908} &	\bf{0.957} \\
			Trouser &	0.807 &	0.803 &	0.971 &	0.971 &	\underline{0.989} &	\underline{0.989} &	0.987 &	\bf{0.998} \\
			Pullover &	0.829 &	0.714 &	0.865 &	0.852 &	0.881 &	\underline{0.893} &	0.877 &	\bf{0.924} \\
			Dress &	0.785 &	0.872 &	0.912 &	0.873 &	0.938 &	0.937 & \underline{0.941} &	\bf{0.971} \\
			Coat &	0.729 &	0.759 &	0.876 &	0.883 &	0.919 &	\underline{0.922} &	0.918 &	\bf{0.944} \\
			Sandal &	\underline{0.931} &	0.927 &	0.896 &	0.871 &	0.911 &	0.922 &	0.910 &	\bf{0.964} \\
			Shirt &	0.667 &	0.810 &	0.743 &	0.734 &	0.836 &	0.836 &	\underline{0.839} &	\bf{0.881} \\
			Sneaker &	0.954 &	0.883 &	0.972 &	0.981 &	\underline{0.987} &	0.985 &	\underline{0.987} &	\bf{0.994} \\
			Bag &	\underline{0.969} &	0.830 &	0.819 &	0.718 &	0.861 &	0.853 &	0.864 &	\bf{0.989} \\
			Ankle &	0.716 &	0.803 &	0.899 &	0.916 &	0.973 &	0.980 &	\underline{0.986} &	\bf{0.992} \\
			\hline
			Avg. &	0.808 &	0.809 &	0.884 &	0.866 &	0.918 &	0.921 &	\underline{0.922} &	\bf{0.961} \\			
			\hline\hline
		\end{tabular}
	\end{center}
	\vspace{-0.25cm} 
	\caption{Results of RSC-VAE in Fashion MNIST dataset. ``Ankle" is the abbreviation of ``Ankle boot".\label{tab:2}}
\end{table}

\begin{table}
	\scriptsize
	\begin{center}
		\linespread{1.05}\selectfont
		\begin{tabular}{L{7.2mm}|L{5mm}L{5mm}L{5mm}L{5mm}|L{5mm}L{5mm}L{5mm}L{5mm}}
			\hline\hline
			Category &	AE &	Gano-maly  &	DSE-BM &	patch-SVDD &	Recode VAE &	RSC-VAE$_o$ &	RSC-VAE$_d$ &	RSC-VAE$_e$ \\
			\hline
			Carpet &	0.641 &	\underline{0.699}  &	0.413 &	0.480 &	0.532 &	0.673 &	0.621 &	\bf{0.745} \\
			Grid &	\underline{0.825} &	0.708  &	0.717 &	0.83 &	0.786 &	0.801 &	0.823 &	\bf{0.846} \\
			Leather &	0.799 &	0.842  &	0.416 &	0.690 &	0.905 &	0.909 &	\underline{0.939} &	\bf{0.982} \\
			Tile &	0.738 &	0.794  &	0.690 &	0.700 &	0.837 &	0.784 & \underline{0.847} &	\bf{0.966} \\
			Wood &	0.970 &	0.834  &	0.952 &	0.940 &	0.948 &	0.950 &	\bf{0.992} & \underline{0.989} \\
			Bottle &	0.654 &	0.892  & 0.818 &	0.920 &	\bf{0.963} &	0.942 &	0.955 &	\underline{0.962} \\
			Cable &	0.639 &	0.757  &	0.685 &	0.570 &	\underline{0.874} &	0.819 &	\bf{0.876} &	0.853 \\
			Capsule &	0.619 &	0.732  &	0.594 &	0.670 &	0.737 &	0.849 &	\underline{0.867} &	\bf{0.880} \\
			Hazelnut &	0.731 &	0.785  &	0.762 &	0.830 &	0.805 &	\underline{0.915} &	0.901 &	\bf{0.934} \\
			Metal &	0.631 &	0.700 &	0.679 &	0.520 &	0.830 &	0.840 & \bf{0.869} & \underline{0.853} \\
			Pill &	0.773 &	0.743  &	0.806 &	0.770 &	0.853 &	0.873 &	\underline{0.902} &	\bf{0.913} \\
			Screw &	\bf{0.999} &	0.746  & \bf{0.999} &	0.560 &	0.974 &	0.977 &	0.975 &	\underline{0.980} \\
			Tooth &	0.769 &	0.653  &	0.781 &	\underline{0.920} & \bf{1.000} & \bf{1.000} &	\bf{1.000} &	\bf{1.000} \\
			Trans &	0.646 &	0.792  &	0.741 &	0.680 &	0.897 &	\underline{0.883} &	\bf{0.893} &	0.880 \\
			Zipper &	0.868 &	0.745  &	0.584 &	0.940 &	0.824 &	\bf{0.868} &	0.853 &	\underline{0.855} \\
			\hline
			Avg. &	0.753 &	0.761  &	0.709 &	0.735 &	0.851 &	0.872 &	\underline{0.888} &	\bf{0.909} \\	
			\hline\hline
		\end{tabular}
	\end{center}
	\vspace{-0.25cm} 
	\caption{Results of RSC-VAE in MVTecAD dataset. ``Metal", ``Tooth", ``Trans" is 
			 the abbreviation of ``Metal\_nut", ``Toothbrush", ``Transistor", respectively.\label{tab:3}}
	\vspace{-0.25cm}
\end{table}

\begin{table}
	\scriptsize
	\begin{center}
		\linespread{1.1}\selectfont
		\begin{tabular}{ccc|ccc} 
			\hline\hline
			Category & VAE &	RSC-VAE$_o$ &	Category &	VAE &	RSC-VAE$_o$  \\
			\hline
			Trouser & 0.333 & \bf{0.229} & Sandal & 0.491 & \bf{0.471} \\	
			\hline	
			T-shirt/tp & 0.390 & \bf{0.251} & Sneaker & \bf{0.248} & 0.261 \\
			\hline
			Pullover  & 0.215 & \bf{0.207} & Shirt & 0.225 & \bf{0.196}\\
			\hline
			Coat & 0.495 & \bf{0.365} & Bag & \bf{0.224} & 0.274\\
			\hline
			Dress & 0.202 & \bf{0.200} & Ankle boot & 0.297 & \bf{0.296} \\
			\hline
			\multicolumn{3}{c|}{} & Average & 0.312 & \bf{0.275} \\
			\hline\hline
		\end{tabular}
	\end{center}
	\vspace{-0.25cm} 
	\caption{Comparison of $S_a$ in fMNIST dataset between VAE and RSC-VAE$_o$.\label{tab:4}}
	\vspace{-0.25cm}
\end{table}

\subsection{Results of RSC-VAE$_o$}
\noindent\textbf{Comparison with baseline and ablation study. }In RSC-VAE$_o$, we train the model with only normal data. Table \ref{tab:1}, \ref{tab:2}, \ref{tab:3} demonstrate the performance of the model on MNIST, fMNIST, and MVTecAD datasets, respectively. We first compare some state-of-the-art models, including Autoencoder, Ganomaly\cite{2018GANomaly} in all three datasets, OCGAN\cite{perera2019ocgan}, GradCon\cite{kwon2020backpropagated} in MNIST, ADGAN\cite{deeckeanomaly}, DSEBM\cite{zhai2016deep} in fMNIST, and DSEBM, PatchSVDD\cite{yi2020patch} in MVTecAD. The proposed algorithm has a significant improvement on either dataset, which illustrates the stability of RSC-VAE in different scenarios. In addition, we performed ablation study on loss function $l_{consist}$. We compare RSC-VAE$_o$ with the baseline model without applying $l_{consist}$ in training stage, namely ``Recode VAE" in above tables. The latter contains the same network structure, hyper-parameters and other configurations with the former. We find that the performance of RSC-VAE$_o$ remained stable on MNIST while improved in the most categories on fMNIST and MVTecAD, where the average AUC increased by 2.1\% and 3.3\%, respectively, fully demonstrating the effectiveness of the $l_{consist}$ term on the recoding based VAE mechanism. 

\vspace{0.05cm}
\noindent\textbf{Verification of corresponding relationship. }
Morever, noting that the proposed RSC-VAE$_o$ reinforces the corresponding relationship between the image space of normal samples and Normal-semantic-region. As a demonstration of it, we divide two component in Eq.\ref{abn-formula} to obtain the indicator $a_i$:

\vspace{-0.2cm}
\begin{sequation}
	a_i = \frac{l_{mut}(x_i)}{\mathbb{E}_x[l_{mut}(x)]} \bigg/  \frac{l_{recon}(x_i)}{\mathbb{E}_x[l_{recon}(x)]}
\end{sequation}

We calculate this indicator for all normal samples in the test set and its standard deviation $S_a$, for the reason that the stability of $a_i$ reflects the strength of the corresponding relationship between two spaces of the normal samples.We compare $S_a$ of each category of fMNIST dataset between VAE and RSC-VAE$_o$ models, shown in Table \ref{tab:4}. It is observed that the latter's $S_a$ decreases under eight categories compared to the former, with 0.036 average decrease.

\subsection{Results of RSC-VAE$_d$}

We illustrate the experimental results in the Distributionally-Shifted Training Mode. Table \ref{tab:1}, \ref{tab:2}, \ref{tab:3} also demonstrate the results of RSC-VAE$_d$ trained on three datasets. Compared with RSC-VAE$_o$, the average AUC of RSC-VAE$_d$ increase by 0.94\%, 0.95\%, and 3.3\% on the three datasets, respectively, which fully demonstrates the enhancement of the model's Semantic-deviation capability when introducing DSA.

\subsection{Results of RSC-VAE$_e$}

\noindent\textbf{Performance compared with the former.} In the Extremely-Imbalanced Training Mode, a few anomalous samples are introduced. For fair comparison, in MNIST/fMNIST dataset, the probability(1\%) of introducing anomalous samples is consistent with that of DSA in RSC-VAE$_d$, while in MVTecAD, we introduce the minimum number of anomaly(1 sample in each anomalous sub-category). As shown in Table \ref{tab:1}, \ref{tab:2}, \ref{tab:3}, the average AUC on three datasets of RSC-VAE respectively increased by 1.8\%, 3.9\%, and 2.1\% when replacing pseudo anomalies with actual ones, demonstrating that the model has captured the actual anomalous information.

\noindent\textbf{Ablation on loss function.} As mentioned in Methodology, the loss function on anomalous sample input consists of two parts, $l_{enc1}$ and $l_{consist}$. Hence we perform ablation experiments to verify the validity of these two parts. We take RSC-VAE$_o$ as the baseline, and demonstrate the results when the loss function for anomalous samples is set to $l_{enc1}$ and $(l_{enc1} + l_{consist})$, respectively. Noting that in the MNIST/fMNIST dataset, we only add 0.5\% proportion of anomalous samples in the original training set. While in MVTecAD, we add only ONE sample in each anomalous sub-category for every anomalous category to the training set. From the experimental results in Table \ref{tab:5}, it is concluded that our results have a remarkable improvement compared with RSC-VAE$_o$ when only applying $l_{enc1}$, and the average AUC increases by 2.0\%, 1.0\%, and 3.3\% in three datasets, respectively, which verifies the mechanism of $l_{enc1}$ to guide the semantics of anomalous samples into the Unknown-semantic-region. Furthermore, after applying $l_{consist}$, the average AUC on three datasets further increased by 0.1\%, 0.9\%, and 0.4\%, respectively, demonstrating that $l_{consist}$ can further reduce the decoder's ability to reconstruct anomalous semantics.

\begin{table}
	\scriptsize
	\begin{center}
		\linespread{1.1}\selectfont
		\begin{tabular}{|c|c|c|c|} 
			\hline\hline
			\multirow{2}{*}{} & \multirow{2}{*}{RSC-VAE$_o$} & \multicolumn{2}{c|}{RSC-VAE$_e$} \\	
			\cline{3-4}
			~ & ~ &	$l_{enc1}$ &	$l_{enc1}$+$l_{consist}$ \\
			\hline
			MNIST &	0.976 &	\underline{0.996} &	\bf{0.997} \\
			\hline
			fMNIST &	0.922 &	\underline{0.944} &	\bf{0.953} \\
			\hline
			MVTecAD &	0.872 &	\underline{0.905} &	\bf{0.909} \\
			\hline\hline
		\end{tabular}
	\end{center}
	\vspace{-0.25cm} 
	
	\caption{Ablation experiment results on loss functions in RSC-VAE$_e$. \label{tab:5}}
	\vspace{-0.35cm}
\end{table}

\begin{figure}[t]
	\begin{center}
		\includegraphics[width=1.03\linewidth]{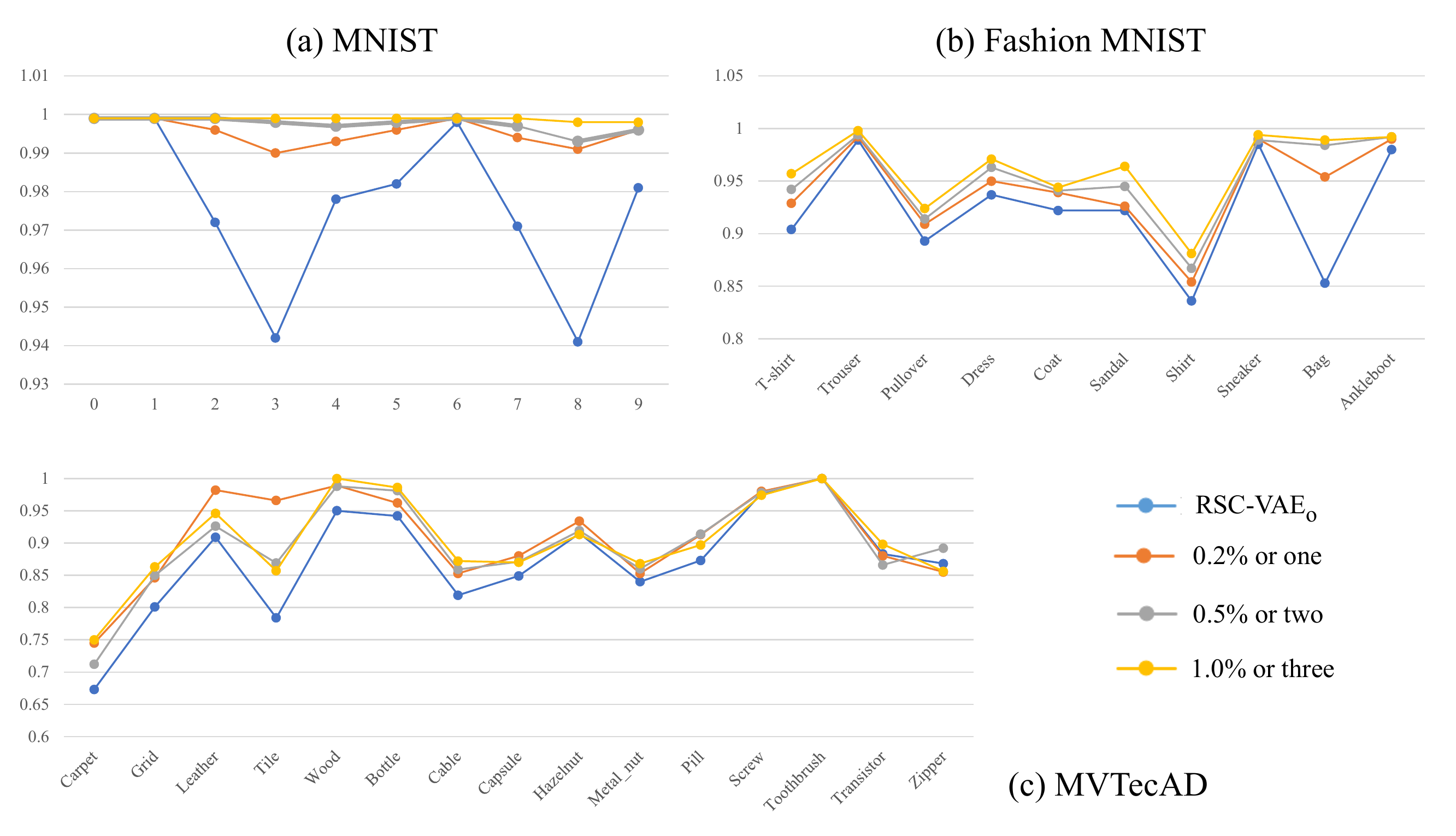}
	\end{center}
	\vspace{-0.4cm} 
	\caption{Results of training with different proportions/numbers of anomalous samples in RSC-VAE$_e$.}
	\label{fig:6}
	\vspace{-0.5cm} 
\end{figure}

\vspace{0.05cm}
\noindent\textbf{Training with different proportions of anomalous samples.} In the ablation study, we added only 0.5\% of original training set anomalous samples to MNIST/fMNIST dataset for training, meaning that the ratio of positive and negative samples is 222:1, which is adequate to demonstrate the RSC-VAE$_e$’s ability to exploit a small number of anomalous samples. Here, we observe the results by changing the proportion of anomalous samples in MNIST/fMNIST to 0.2\% and 1\%; Similarly, in MVTecAD, we increase the number of randomly selected anomalous samples to 2 and 3. The results are shown in Figure \ref{fig:6}. It can be clearly found that as the anomaly information is continuously introduced in the training set, model’s Semantic-deviation ability for the anomalous input strengthens.

\begin{figure}[t]
	\centering
	\includegraphics[width=1.05\linewidth]{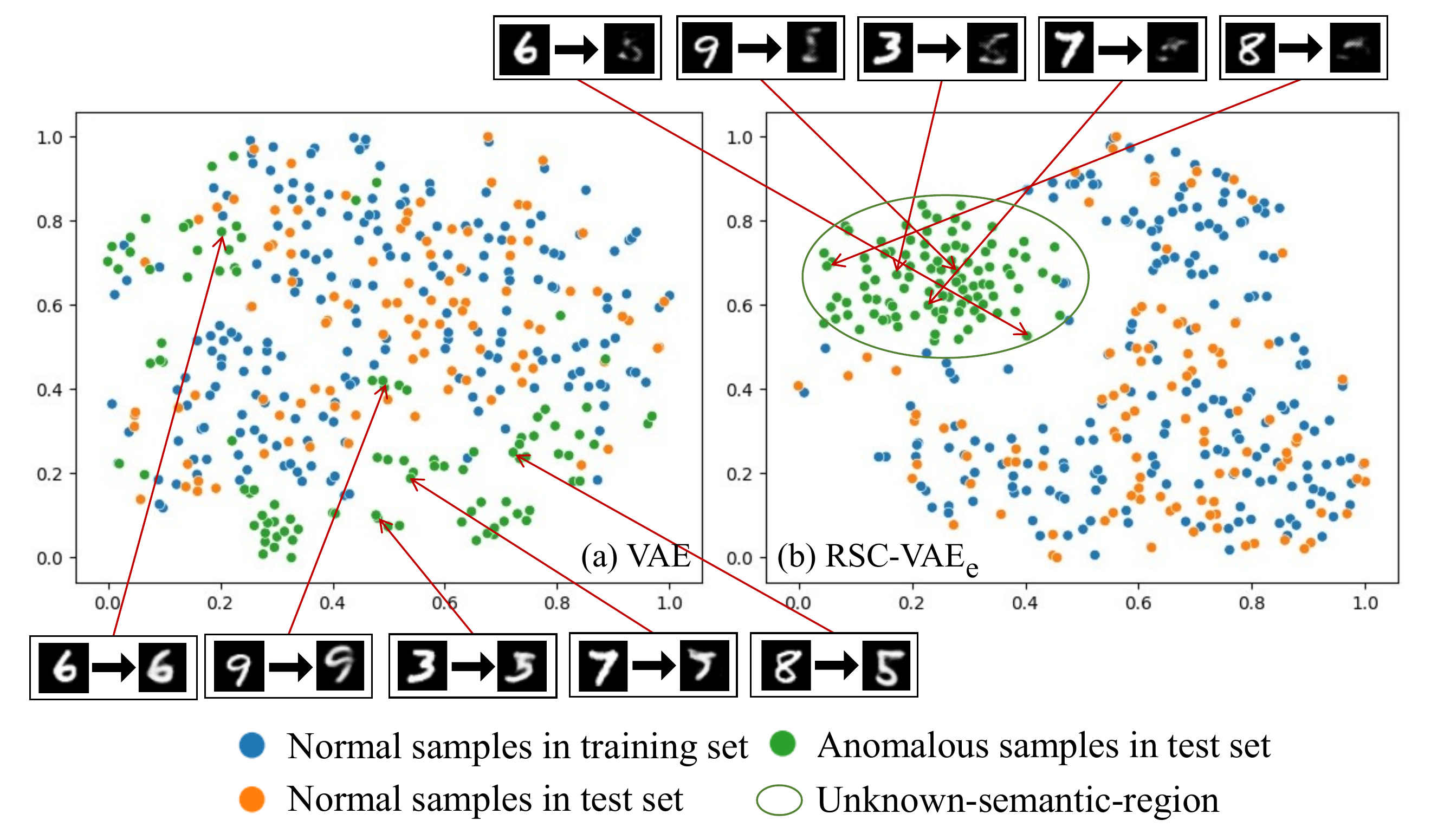}
	\caption{t-SNE visualization of Unknown-semantic-region and Semantic-deviation. We take five different digits in MNIST as example, putting them into VAE and RSC-VAE$_e$ model, where the normal category is digit ``5”, respectively, then we can compare the latent space and the reconstruction results of them.}
	\label{fig:7}
	\vspace{-0.5cm} 
\end{figure}

\noindent\textbf{Verification of Semantic-deviation. }We use the converged VAE model with the digit ``5" in the MNIST dataset as the training set and the RSC-VAE$_e$ model with the same setting and structure. We encode the normal samples from the training set as well as the normal and the anomalous samples from the test set. We adopt t-SNE\cite{maaten2008visualizing} to reduce the dimension of vector $\mu$ in the encoding results to 2 and demonstrate the encoding and reconstruction results of five different anomalous inputs. As shown in Figure \ref{fig:7}, VAE cannot control the encoding results of samples, leading to the mixture of the feature representation from normal and anomalous samples. Morever, the input can be well reconstructed (digit ``6”, ``9”) or reconstructed as a normal category (digit ``3”, ``7”, ``8”), which the illustrates the decoder’s unstable ``generalization" ability; While in RSC-VAE$_e$, because of the Semantic-deviation mechanism, the feature representations of anomalous samples are clustered in Unknown-semantic-region that lack meaningful information, hence leading to degraded reconstructions.

\section{Conclusion}
In this paper, we conduct an in-depth study around the value of VAE’s latent space. Firstly, three different types of latent space region, Normal/Anomalous/Unknown-Semantic-Regions, are proposed. On the basis of this, we combine VAE with recoding mechanism and propose Recoding Semantic Consistency Based VAE (RSC-VAE), which constrains the consistency of the semantics of two encodings, and propose three different training modes for this model. We hope our work can provide beneficial insight for subsequent anomaly detection tasks.


\newpage

{\small
\bibliographystyle{ieee_fullname}
\bibliography{egbib}

\begin{thebibliography}{10}\itemsep=-1pt

\bibitem{2018GANomaly}
Samet Akcay, Amir Atapour-Abarghouei, and Toby~P Breckon.
\newblock Ganomaly: Semi-supervised anomaly detection via adversarial training.
\newblock 2018.

\bibitem{8294186}
N. {Akhtar} and A. {Mian}.
\newblock Threat of adversarial attacks on deep learning in computer vision: A
  survey.
\newblock {\em IEEE Access}, 6:14410--14430, 2018.

\bibitem{bergman2020classification}
Liron Bergman and Yedid Hoshen.
\newblock Classification-based anomaly detection for general data.
\newblock {\em arXiv preprint arXiv:2005.02359}, 2020.

\bibitem{bergmann2019mvtec}
Paul Bergmann, Michael Fauser, David Sattlegger, and Carsten Steger.
\newblock Mvtec ad--a comprehensive real-world dataset for unsupervised anomaly
  detection.
\newblock In {\em Proceedings of the IEEE Conference on Computer Vision and
  Pattern Recognition}, pages 9592--9600, 2019.

\bibitem{chen2020simple}
Ting Chen, Simon Kornblith, Mohammad Norouzi, and Geoffrey Hinton.
\newblock A simple framework for contrastive learning of visual
  representations.
\newblock In {\em International conference on machine learning}, pages
  1597--1607. PMLR, 2020.

\bibitem{deeckeanomaly}
Lucas Deecke, Robert Vandermeulen, Lukas Ruff, Stephan Mandt, and Marius Kloft.
\newblock Anomaly detection with generative adversarial networks, 2018.
\newblock In {\em URL https://openreview. net/forum}.

\bibitem{devries2017improved}
Terrance DeVries and Graham~W Taylor.
\newblock Improved regularization of convolutional neural networks with cutout.
\newblock {\em arXiv preprint arXiv:1708.04552}, 2017.

\bibitem{fei2020attribute}
Ye Fei, Chaoqin Huang, Cao Jinkun, Maosen Li, Ya Zhang, and Cewu Lu.
\newblock Attribute restoration framework for anomaly detection.
\newblock {\em IEEE Transactions on Multimedia}, 2020.

\bibitem{golan2018deep}
Izhak Golan and Ran El-Yaniv.
\newblock Deep anomaly detection using geometric transformations.
\newblock In {\em Advances in Neural Information Processing Systems}, pages
  9758--9769, 2018.

\bibitem{2020Memorizing}
Dong Gong, Lingqiao Liu, Vuong Le, Budhaditya Saha, Moussa~Reda Mansour, Svetha
  Venkatesh, and Anton Van~Den Hengel.
\newblock Memorizing normality to detect anomaly: Memory-augmented deep
  autoencoder for unsupervised anomaly detection.
\newblock In {\em 2019 IEEE/CVF International Conference on Computer Vision
  (ICCV)}, 2020.

\bibitem{goodfellow2014generative}
Ian Goodfellow, Jean Pouget-Abadie, Mehdi Mirza, Bing Xu, David Warde-Farley,
  Sherjil Ozair, Aaron Courville, and Yoshua Bengio.
\newblock Generative adversarial nets.
\newblock In {\em Advances in neural information processing systems}, pages
  2672--2680, 2014.

\bibitem{he2020momentum}
Kaiming He, Haoqi Fan, Yuxin Wu, Saining Xie, and Ross Girshick.
\newblock Momentum contrast for unsupervised visual representation learning.
\newblock In {\em Proceedings of the IEEE/CVF Conference on Computer Vision and
  Pattern Recognition}, pages 9729--9738, 2020.

\bibitem{hong2020latent}
Eungi Hong and Yoonsik Choe.
\newblock Latent feature decentralization loss for one-class anomaly detection.
\newblock {\em IEEE Access}, 8:165658--165669, 2020.

\bibitem{hou2017deep}
Xianxu Hou, Linlin Shen, Ke Sun, and Guoping Qiu.
\newblock Deep feature consistent variational autoencoder.
\newblock In {\em 2017 IEEE Winter Conference on Applications of Computer
  Vision (WACV)}, pages 1133--1141. IEEE, 2017.

\bibitem{jha2018disentangling}
Ananya~Harsh Jha, Saket Anand, Maneesh Singh, and VSR Veeravasarapu.
\newblock Disentangling factors of variation with cycle-consistent variational
  auto-encoders.
\newblock In {\em Proceedings of the European Conference on Computer Vision
  (ECCV)}, pages 805--820, 2018.

\bibitem{kim2019rapp}
Ki~Hyun Kim, Sangwoo Shim, Yongsub Lim, Jongseob Jeon, Jeongwoo Choi, Byungchan
  Kim, and Andre~S Yoon.
\newblock Rapp: Novelty detection with reconstruction along projection pathway.
\newblock In {\em International Conference on Learning Representations}, 2019.

\bibitem{kingma2013auto}
Diederik~P Kingma and Max Welling.
\newblock Auto-encoding variational bayes.
\newblock {\em arXiv preprint arXiv:1312.6114}, 2013.

\bibitem{komoto2020consistency}
Kyosuke Komoto, Hiroaki Aizawa, and Kunihito Kato.
\newblock Consistency ensured bi-directional gan for anomaly detection.
\newblock In {\em International Workshop on Frontiers of Computer Vision},
  pages 236--247. Springer, 2020.

\bibitem{kullback1951}
S. Kullback and R.~A. Leibler.
\newblock On information and sufficiency.
\newblock {\em Ann. Math. Statist.}, 22(1):79--86, 03 1951.

\bibitem{kwon2020backpropagated}
Gukyeong Kwon, Mohit Prabhushankar, Dogancan Temel, and Ghassan AlRegib.
\newblock Backpropagated gradient representations for anomaly detection.
\newblock In {\em European Conference on Computer Vision}, pages 206--226.
  Springer, 2020.

\bibitem{lecun1998mnist}
Yann LeCun.
\newblock The mnist database of handwritten digits.
\newblock {\em http://yann. lecun. com/exdb/mnist/}.

\bibitem{lecun2015deep}
Yann LeCun, Yoshua Bengio, and Geoffrey Hinton.
\newblock Deep learning.
\newblock {\em nature}, 521(7553):436--444, 2015.

\bibitem{liznerski2020explainable}
Philipp Liznerski, Lukas Ruff, Robert~A Vandermeulen, Billy~Joe Franks, Marius
  Kloft, and Klaus-Robert M{\"u}ller.
\newblock Explainable deep one-class classification.
\newblock {\em arXiv preprint arXiv:2007.01760}, 2020.

\bibitem{maaten2008visualizing}
Laurens van~der Maaten and Geoffrey Hinton.
\newblock Visualizing data using t-sne.
\newblock {\em Journal of machine learning research}, 9(Nov):2579--2605, 2008.

\bibitem{makhzani2015adversarial}
Alireza Makhzani, Jonathon Shlens, Navdeep Jaitly, Ian Goodfellow, and Brendan
  Frey.
\newblock Adversarial autoencoders.
\newblock {\em arXiv preprint arXiv:1511.05644}, 2015.

\bibitem{paszke2017automatic}
Adam Paszke, Sam Gross, Soumith Chintala, Gregory Chanan, Edward Yang, Zachary
  DeVito, Zeming Lin, Alban Desmaison, Luca Antiga, and Adam Lerer.
\newblock Automatic differentiation in pytorch.
\newblock 2017.

\bibitem{perera2019ocgan}
Pramuditha Perera, Ramesh Nallapati, and Bing Xiang.
\newblock Ocgan: One-class novelty detection using gans with constrained latent
  representations.
\newblock In {\em Proceedings of the IEEE Conference on Computer Vision and
  Pattern Recognition}, pages 2898--2906, 2019.

\bibitem{ruff2020unifying}
Lukas Ruff, Jacob~R Kauffmann, Robert~A Vandermeulen, Gr{\'e}goire Montavon,
  Wojciech Samek, Marius Kloft, Thomas~G Dietterich, and Klaus-Robert
  M{\"u}ller.
\newblock A unifying review of deep and shallow anomaly detection.
\newblock {\em arXiv preprint arXiv:2009.11732}, 2020.

\bibitem{ruff2018deep}
Lukas Ruff, Robert Vandermeulen, Nico Goernitz, Lucas Deecke, Shoaib~Ahmed
  Siddiqui, Alexander Binder, Emmanuel M{\"u}ller, and Marius Kloft.
\newblock Deep one-class classification.
\newblock In {\em International conference on machine learning}, pages
  4393--4402, 2018.

\bibitem{schlegl2017unsupervised}
Thomas Schlegl, Philipp Seeb{\"o}ck, Sebastian~M Waldstein, Ursula
  Schmidt-Erfurth, and Georg Langs.
\newblock Unsupervised anomaly detection with generative adversarial networks
  to guide marker discovery.
\newblock In {\em International conference on information processing in medical
  imaging}, pages 146--157. Springer, 2017.

\bibitem{selvaraju2017grad}
Ramprasaath~R Selvaraju, Michael Cogswell, Abhishek Das, Ramakrishna Vedantam,
  Devi Parikh, and Dhruv Batra.
\newblock Grad-cam: Visual explanations from deep networks via gradient-based
  localization.
\newblock In {\em Proceedings of the IEEE international conference on computer
  vision}, pages 618--626, 2017.

\bibitem{6773024}
C.~E. {Shannon}.
\newblock A mathematical theory of communication.
\newblock {\em The Bell System Technical Journal}, 27(3):379--423, 1948.

\bibitem{tack2020csi}
Jihoon Tack, Sangwoo Mo, Jongheon Jeong, and Jinwoo Shin.
\newblock Csi: Novelty detection via contrastive learning on distributionally
  shifted instances.
\newblock In {\em 34th Conference on Neural Information Processing Systems
  (NeurIPS) 2020}. Neural Information Processing Systems, 2020.

\bibitem{ulutas2020split}
Tolga Ulutas, Muhammed Ali~Nur Oz, Muharrem Mercimek, and Ozgur~Turay Kaymakci.
\newblock Split-brain autoencoder approach for surface defect detection.
\newblock In {\em 2020 International Conference on Electrical, Communication,
  and Computer Engineering (ICECCE)}, pages 1--5. IEEE, 2020.

\bibitem{venkataramanan2019attention}
Shashanka Venkataramanan, Kuan-Chuan Peng, Rajat~Vikram Singh, and Abhijit
  Mahalanobis.
\newblock Attention guided anomaly detection and localization in images.
\newblock {\em arXiv preprint arXiv:1911.08616}, 2019.

\bibitem{xiao2017fashion}
Han Xiao, Kashif Rasul, and Roland Vollgraf.
\newblock Fashion-mnist: a novel image dataset for benchmarking machine
  learning algorithms.
\newblock {\em arXiv preprint arXiv:1708.07747}, 2017.

\bibitem{yi2020patch}
Jihun Yi and Sungroh Yoon.
\newblock Patch svdd: Patch-level svdd for anomaly detection and segmentation.
\newblock In {\em Proceedings of the Asian Conference on Computer Vision},
  2020.

\bibitem{zhai2016deep}
Shuangfei Zhai, Yu Cheng, Weining Lu, and Zhongfei Zhang.
\newblock Deep structured energy based models for anomaly detection.
\newblock In {\em International Conference on Machine Learning}, pages
  1100--1109, 2016.

\bibitem{zhao2018surface}
Zhixuan Zhao, Bo Li, Rong Dong, and Peng Zhao.
\newblock A surface defect detection method based on positive samples.
\newblock In {\em Pacific Rim International Conference on Artificial
  Intelligence}, pages 473--481. Springer, 2018.

\end{thebibliography}
}

\end{document}